\documentclass[runningheads]{llncs}
\usepackage{graphicx}
\usepackage{amssymb}
\usepackage{amsmath}
\usepackage{multirow}
\usepackage{booktabs}
\usepackage{textcomp}
\usepackage{adjustbox}
\usepackage{array}

\begin{document}
\title{SOE: SO(3)-Equivariant 3D MRI Encoding}
\titlerunning{SOE: SO(3)-Equivariant 3D MRI Encoding}

\author{Shizhe He\inst{1} \and
Magdalini Paschali\inst{2} \and
Jiahong Ouyang\inst{3}  \and Adnan Masood \inst{4} \and \\
Akshay Chaudhari\inst{2, 5} \and
Ehsan Adeli\inst{1, 3}\thanks{Corresponding author: \texttt{eadeli@stanford.edu}}}

\authorrunning{S. He et al.}

\institute{Department of Computer Science, Stanford University, Stanford, CA, USA \and
Department of Radiology, Stanford University, Stanford, CA, USA \and
Department of Psychiatry and Behavioral Sciences, Stanford University, \\Stanford, CA, USA \and
UST, Aliso Viejo, CA, USA
\and
Department of Biomedical Data Science, Stanford University, Stanford, CA, USA
}

\maketitle              
\begin{abstract}

Representation learning has become increasingly important, especially as powerful models have shifted towards learning latent representations before fine-tuning for downstream tasks. This approach is particularly valuable in leveraging the structural information within brain anatomy. However, a common limitation of recent models developed for MRIs is their tendency to ignore or remove geometric information, such as translation and rotation, thereby creating invariance with respect to geometric operations. We contend that incorporating knowledge about these geometric transformations into the model can significantly enhance its ability to learn more detailed anatomical information within brain structures. As a result, we propose a novel method for encoding 3D MRIs that enforces equivariance with respect to all rotations in 3D space, in other words, SO(3)-equivariance (SOE). By explicitly modeling this geometric equivariance in the representation space, we ensure that any rotational operation applied to the input image space is also reflected in the embedding representation space. This approach requires moving beyond traditional representation learning methods, as we need a representation vector space that allows for the application of the same SO(3) operation in that space. To facilitate this, we leverage the concept of vector neurons. The representation space formed by our method, SOE, captures the brain's structural and anatomical information more effectively.
We evaluate SOE pretrained on the structural MRIs of two public data sets with respect to the downstream task of predicting age and diagnosing Alzheimer's Disease from T1-weighted brain scans of the ADNI data set. We demonstrate that our approach not only outperforms other methods but is also robust against various degrees of rotation along different axes. The code is available at \texttt{https://github.com/shizhehe/SOE-representation-learning}.

\keywords{Contrastive Learning \and SO(3)-Equivariance \and 3D Brain MRI.}
\end{abstract}
\section{Introduction}
In recent years, there has been a significant surge in the development and adoption of various self-supervised representation learning techniques \cite{SimCLR_model,BYOL_model,HomE_model,CPC_model,MoCo_model}. These methods aim to create universal approaches (a.k.a. foundation models) for extracting meaningful features or representations from unlabeled data. This approach has gained popularity due to its ability to utilize large amounts of readily available unlabeled data, reducing the dependency on costly and time-consuming labeling processes. Various pretraining techniques have recently also been developed for medical images
\cite{huang2023self,Jiahong_longitudinal,spherical_cnns}. The recent methods developed for medical images, including magnetic resonance imaging (MRI), have mostly replicated pretraining techniques used in natural images applications. Examples of such are MRI masked autoencoders \cite{lang20233d}. 

A prevalent issue with recent models designed for MRI data is their tendency to overlook or eliminate geometric information, such as translation and rotation (e.g., \cite{esfahani2020compressed,kwon2023rotation}), resulting in invariance to geometric transformations. We argue that making MRI encoders understand these geometric transformations can improve their capacity to capture intricate anatomical details within brain structures. 

In this paper, we introduce a novel approach for encoding 3D MRIs that enforces SO(3)-equivariance (SOE). 

SO(3) refers to the special orthogonal group in three dimensions, which represents all possible rotations in 3D space \cite{spherical_cnns}. Mathematically, it consists of all $3\times3$ orthogonal matrices with a determinant of $+1$. Traditionally, in computer vision, graphics, robotics, and physics, these matrices describe the rotation of an object in 3D space without any reflection or scaling. SO(3) rotations are commonly used to represent the orientation of objects or coordinate systems \cite{vectorneurons,steerable_filters,spherical_cnns}. 
By explicitly incorporating geometric equivariance, we ensure that any operation performed on the input image space is correspondingly mirrored in the embedding representation space. This strategy necessitates a departure from conventional representation learning methods, as it requires a representation vector space (or vector neurons) that accommodates the application of the same SO(3) operation in that space. 

In summary, our contributions are: (1) We propose a method to learn {geo\-metry}-aware representations by modeling rotation through SO(3) matrix transformation and enforcing SO(3)-equivariance between the original 3D MRI space with the latent representation space (see Fig.~\ref{encoder_architecture}). Here, we introduce a technique to regularize the model to prevent it from falling into trivial solutions that completely ignore SO(3) operations. (2) We formulate this self-supervised learning framework within a model architecture composed of a generic encoder, the Vector Neuron module \cite{vectorneurons}, and a prediction head. (3) We show that incorporating geometric information during pretraining improves the performance on AD classification and age prediction tasks.

\section{Methods}
Here, we present SO(3)-Equivariant (SOE) Representation Learning for 3D MRI encoding. Let $Rot(\cdot)$ be a 3D rotation function that maps a 3D medical scan to its rotated orientation and $x_i^1$ be a 3D MRI input whose representation is generated using the encoder network $f_{\theta}$. 
Given an arbitrary rotated version of $x_i^1$ by $x_i^2 = Rot(x_i^1)$, we argue that optimizing for rotational equivariance between the representations $f_{\theta}(x_i^1) = z_i^1$ and $f_{\theta}(x_i^2) = z_i^2$, the encoder learns structural coherence in the 3D MRI. To enforce this equivariance, we create a self-supervised setup, in which the same $Rot(\cdot)$ transformation applies to both spaces, i.e., the MRI input and the latent representation. To enable this, we need to turn the latent space into a vector space to enable the application of matrix transformation operations. To this end, we supplement our generic encoder with a Vector Neuron (VN) module \cite{vectorneurons}. 

First, we describe our approach to modeling 3D image rotation $Rot(\cdot)$ as matrix transformations of the SO(3) group in Section \ref{image_rotation}. We then define our proposed pretext representation learning objective 
and offer a more detailed insight into the VN module \cite{vectorneurons}. 

Next, we propose an inverse maximization regularizer for the pretext training objective to prevent the representationsfrom converging to the same point in feature space and avoid trivial solutions. Finally, we introduce a robustness regularization term for downstream task training.

\subsection{Modeling 3D Rotation as SO(3) Matrix Transformations} \label{image_rotation}
 
Let $R_{rot}$ be a $3\times3$ rotation matrix defined by an axis and rotation angle. There exists no straight-forward method (such as matrix multiplication) of applying $R_{rot}$ on a three-dimensional image volume $x^1$ of shape $n \times n \times n$ in non-coordinate representation, where $n$ is the dimensionality of the 3D volume. We define $Rot(\cdot)$ (Fig. \ref{encoder_architecture}) as follows: (1) mapping $x^1$ to its coordinate representation $y^1$ of shape $n \times n \times n \times 3$ where the entries $y^1_{(i, j, k, c)}$ are centered around the origin of the coordinate system, (2) applying $R_{rot}$ to the spatial coordinate representation of each pixel $y^1_{(i, j, k, c)}$ and thereby computing the new rotated location $(p_1, p_2, p_3)$ of each voxel within the image grid $y^2_{(i, j, k, c)}$, (3) mapping and clamping the spatial coordinate representation $y^2$ into the image grid $x^2$, and (4) performing trilinear interpolation for pixel coordinates $(p_1, p_2, p_3)$ that do not fit into exact entries of the image grid. Since $R_{rot}$ is defined by the axis and angle of the applied rotation, we omit them from the definition of $Rot(\cdot)$ for simplicity. Therefore, the rotated volume $x^2$ is defined as:

\begin{equation} \label{rot_equation}
    x^2_{(i, j, k, c)} = Rot(x^1_{(i, j, k, c)}, R_{rot}) = \text{Interpolation}\left(R_{rot} x^1_{(i, j, k, c)} \right).
\end{equation}

\subsection{SO(3)-Equivariant Representation Learning} \label{pretext_definition}
\begin{figure}[t]
\centering
\includegraphics[width=0.8\textwidth, angle=0]{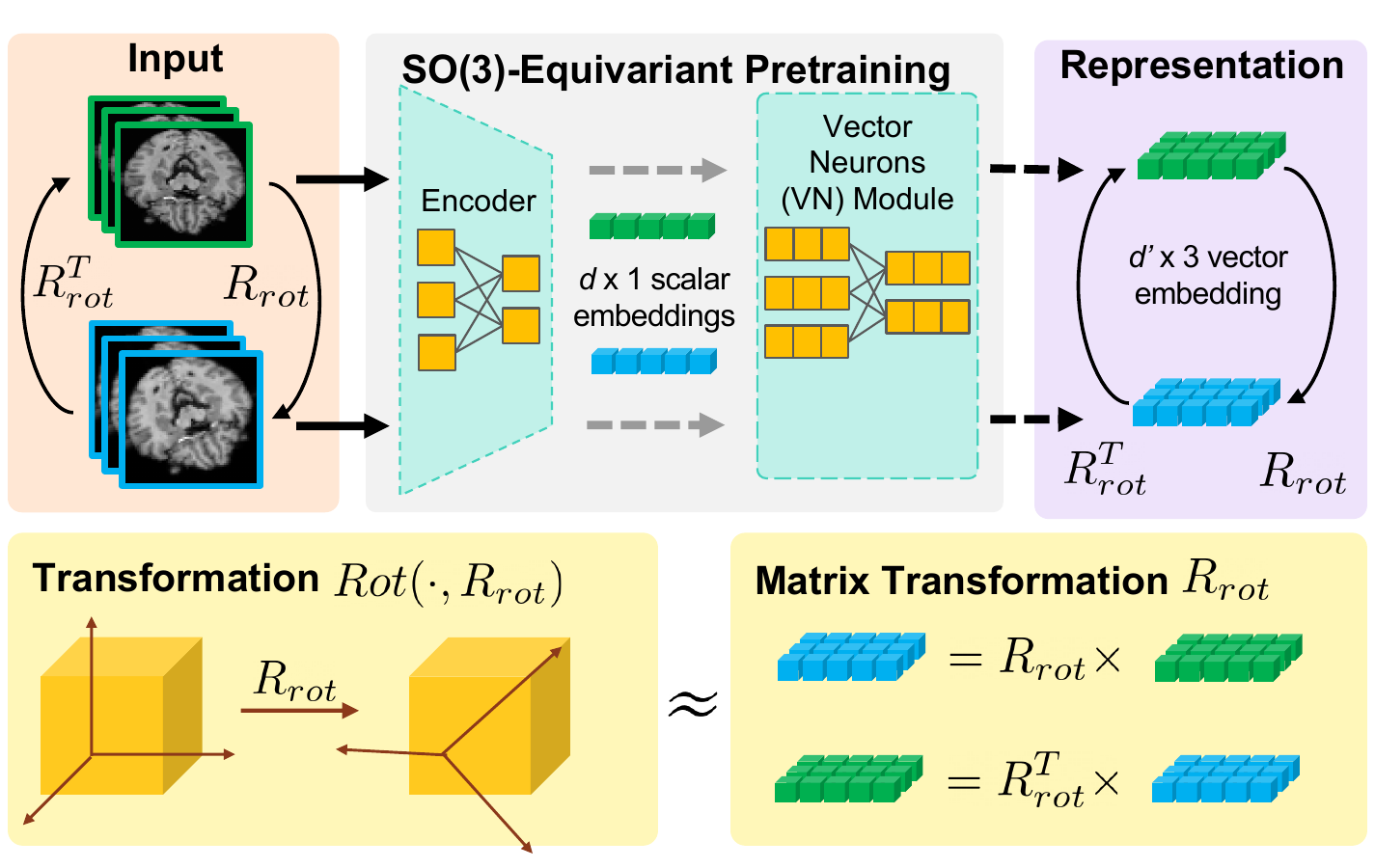}
\caption{SOE Model Architecture: The model takes in a pair of unrotated and rotated samples $x^1, x^2$ as input and maps them to representations preserving the rotational transformation $R_{rot}$. The Encoder encodes the input volume into a $d$ dimensional scalar array and the VN module maps that scalar array into a $d' \times 3$ dimensional vector embedding. $R_{rot}$ is applied on the input volume through the $Rot(\cdot)$ function.} \label{encoder_architecture}
\end{figure}

We define a pretext training objective to learn representations that optimize towards SO(3)-equivariance of the 3D MRIs and their feature representations. To do so, we define our feature encoder (Fig.~\ref{encoder_architecture}) $f(\cdot)$ to map the input $n_{vol} \times n_{vol} \times n_{vol}$ 3D volume to the corresponding $n_{hidden} \times 1$ dimensional representation. Furthermore, we define our Vector Neuron module (VN), which maps the $n_{hidden} \times 1$ dimensional scalar representation to a $n_{hidden} \times 3$ vector representation (details in Section \ref{vn_module}). With $x^2$ being the rotated version of $x^1$, we define the self-supervised representation learning objective to be
\begin{equation} \label{pretext_loss}
    \mathcal{L}_{SO(3)}
    = \lVert \text{VN}(f(x^1))R_{rot} - \text{VN}(f(x^2))\rVert^2 + \lVert \text{VN}(f(x^1)) - \text{VN}(f(x^2))R^T_{rot}\rVert^2.
\end{equation}

This loss function enforces rotational equivariance in the representation space by minimizing the Frobenius distance between $VN(f(x^2))$, the representation of the rotated input sample $x^2$, and $VN(f(x^1))R_{rot}$, the representation of the unrotated input sample transformed to approximate the representation of $x^2$. We model the representations to preserve the symmetry of SO(3)-matrix/rotational transformations from image to feature space. 

\subsection{Vector Neuron Module} \label{vn_module}
The Vector Neuron (VN) module (see Fig.~\ref{encoder_architecture}, middle) plays a pivotal role in our self-supervised learning task as described in Section \ref{pretext_definition}. We use VN \cite{vectorneurons} to map $d \times 1$ scalar representations to $d' \times 3$ vector representations of the input samples where $d$ is the output dimension of the encoder. In other words, the VN layers project the conventional scalar neurons into 3D vector neurons to enable enforcing full SO(3)-equivariance between the input and the representation space. This way, we can use vector representation as 3D points to compute our self-supervised equivariance loss $\mathcal{L}_{SO(3)}$ (Eq. \eqref{pretext_loss}). 

\subsection{Inverse Maximization Component} \label{inverse_component}

The proposed equivariance loss $\mathcal{L}_{SO(3)}$ (Eq. \eqref{pretext_loss}) may encourage feature collapse, where the model learns to map all input to a very small region in the feature space, since $\mathcal{L}_{SO(3)}$ can be naively optimized as inputs converge to the same point in feature space. This would hinder our representation learning process.
To avoid such trivial cases, we propose a regularization component to inversely penalize decreasing distance between the representations of $x^1$ and $x^2$:

\begin{equation} \label{combined_loss}
    \mathcal{L}_{comb}(x^1, x^2) = \mathcal{L}_{SO(3)} + \lambda  \frac{1}{\lVert f(x^1) - f(x^2)\rVert^2},
\end{equation}
where $\lambda$ controls the strength of this regularization. This way, we intuitively encourage diversity in the embedding space and prevent all instances in the feature space from converging to the same point in space.

\subsection{Robustness Regularizer against Rotations} \label{model_robustness}
One of the main benefits of SOE is the robustness against 3D rotations. Here, we define the notion of robustness regularization during the downstream training of our model. We need to explicitly incorporate this aspect beyond merely modeling SO(3)-equivariance in the representation space during pretext training since traditional downstream loss functions do not learn/preserve robustness against geometric transformations by themselves \cite{ilyas2019adversarial}. Rotational robustness is when a trained model is able to consistently predict the correct label/age under all degrees of rotations. We define robustness in this context as directly incorporating robustness modeled as SO(3)-invariance into the downstream objective function as a regularizer. In other words, the terms introduced in Eq. \eqref{pretext_loss} will be added to the downstream fine-tuning objective. 

\section{Experimental Results} \label{experimental_setup}
We design various experiments to examine the performance and generalizability of our representation learning approach.

\noindent\textbf{Datasets.}
We evaluate our approach using two 3D brain MRI datasets for pretraining with representation learning and one dataset for pretraining and subsequent fine-tuning on two downstream tasks. The Alzheimer’s Disease Neuroimaging Initiative (ADNI) dataset~\cite{ADNI} consists of 2,577 T1-weighted MRIs of 811 subjects, where each subject has two to six visits. The first downstream task on ADNI is disease classification into four classes: cognitively normal (NC) (N=214), Alzheimer's disease (AD) (N=187), static MCI (sMCI) (N=275), and progressive MCI (pMCI) (N=135). The subject age ranges between 54.4 and 90.9 (mean 75.18$\pm$6.84). The second downstream task on ADNI is age regression.

The NCANDA study~\cite{pohl2021ncandarelease} recruited 831 participants across five sites in the United States and performed yearly imaging assessments~\cite{brown2015national}. In this work, we used 3,830 T1-weighted MRIs of 831 subjects. The NCANDA dataset was used solely for pretext representation learning. We downsample both datasets to a resolution of $64 \times 64 \times 64$~\cite{Jiahong_longitudinal}. We split the datasets into training (70\%), validation (10\%), and testing (20\%) sets and into five folds for stratified cross-validation by subject~\cite{Jiahong_longitudinal}.

\noindent\textbf{Implementation.}
All self-supervised pretraining approaches are based on a convolutional encoder backbone composed of four convolution blocks, amounting to $97,584$ trainable parameters. Each of the convolution blocks consists of 3D Conv (kernel size $3 \times 3 \times 3$), 3D BatchNorm, LeakyReLU/ReLU (slope $0.2$), 3D Dropout, and 3D MaxPool (kernel size $2$) layers built on PyTorch version 2.0.1~\cite{pytorch}. All models were trained on an NVIDIA GeForce RTX 2080 Ti GPU. Note that we also examined newer transformer-based models, namely the 3D SWIN Transformer \cite{SWIN} and Vision Transformer \cite{ViT} as the backbone encoder in SOE. However, the results show no significant increase against the convolutional encoder backbone. Hence, for the sake of simplicity and consistency, we make all comparisons using conventional backbones. 

During pretraining for both SOE and baseline models, the mini-batch size is set to either 64 or 32. For the downstream tasks, the mini-batch size is set to 64, All models are trained for a maximum of 50 epochs for both pretext and downstream training. The learning rates are initialized to a value varying between $0.01$ and $0.0001$, and decrease in logarithmic steps during training.

\noindent\textbf{Baselines.}
To examine the performance of our proposed approach, we compare our model against other representation learning approaches based on the downstream metrics. For the classification tasks we report balanced accuracy (BACC), F1 score and their standard deviation across folds. For regression we measure coefficient of determination (R2), and mean absolute error (MAE). 

We use the following baseline models as a comparison to SOE: (1) Autoencoder (AE), which learns efficient low-dimensional representations; (2) Variational Autoencoder (VAE), which learns a continuous latent space for its input data \cite{VAE_model}; (3) Masked Autoencoder (MAE), a variant of AE where random patches of the input are masked during training \cite{MAE_model}; and (4) SimCLR, a contrastive self-supervised learning framework \cite{SimCLR_model}.

\noindent\textbf{Robustness Experiments.}
We perform rotational robustness experiments comparing SOE to the baseline model with no pretraining on different levels of rotation. Both models are trained with the robustness regularizer to incorporate rotational robustness in downstream task optimization. We evaluate the models' performance against (1) mild rotations chosen randomly between 15 and 45 degrees and (2) rotations in 90 degree intervals from 0. Note that rotations not in 90 degree intervals from 0 will experience interpolation as described in Section \ref{image_rotation}. We train the robustness models on the two levels of rotation and test them on no rotation and the respective rotations. 

\subsection{Pretraining and Dataset Impact} \label{pretext}
First, we evaluated our proposed method with regard to its generalizability across datasets. We examined the classification and age regression performance on ADNI of models without pretraining, pretrained with SOE on the NCANDA dataset, and pretrained with SOE on the ADNI dataset. In Table \ref{NCANDA_pretraining}, we observe that our proposed SOE pretraining performs better than the no pretraining instance for both ADNI NC vs. AD classification and age prediction tasks by $0.5$-$8\%$ in terms of BACC. Furthermore, we observe that pretraining our model on the same dataset (ADNI) as the downstream task evaluation, the so-called ``self-training", outperforms the model pretrained on NCANDA in classification (AUC 91.3 vs. 91.1; BACC  83.84 vs. 82.81; F1 79.38 vs. 78.50). Particularly when evaluated on the task of age prediction, our representation learning framework outperforms the model with no pretraining (R2 0.45 vs. 0.41; MAE 5.88 vs. 5.95).
Consequently, this confirms our expectation that SOE representation learning allows the model to extract useful information and map to expressive representations. We decide to continue with the ``self-training" setup, i.e., self-supervisd pretraining and downstream training on ADNI, for all further experiments.

\begin{table}[t]\label{NCANDA_pretraining}
\centering
\caption{NCANDA vs. ADNI pretraining, downstream task evaluated on ADNI. (a) NC vs. AD classification cross-validated; (b) NC age regression cross-validated. Our model (SOE) outperforms all other models across both tasks.} \vspace{-8pt}
\resizebox{0.48\textwidth}{!}{
\begin{tabular}{w{l}{2cm}w{c}{2.3cm}w{c}{2.3cm}}
\multicolumn{3}{c}{(a) Classification} \\
  \toprule
Pretraining & BACC\textsuperscript{$\uparrow$}  & F1\textsuperscript{$\uparrow$}               \\ \midrule
 -                             & 82.34 $\pm$ 4.24           & 78.15 $\pm$ 5.56            \\
 NCANDA                          & 82.81 $\pm$  2.01         & 78.50  $\pm$  2.92          \\
 ADNI                            & \textbf{83.84 $\pm$ 4.11}              & \textbf{79.38 $\pm$ 5.58}            \\
 \bottomrule
 \end{tabular}
 } ~
  \resizebox{0.48\textwidth}{!}{
 \begin{tabular}{w{l}{2cm}w{c}{2.3cm}w{c}{2.3cm}}
\multicolumn{3}{c}{(b) Age Regression} \\
\toprule
Pretraining & R2\textsuperscript{$\uparrow$}   & MAE (years)\textsuperscript{$\downarrow$}            \\ \midrule
-                                  & 0.36 $\pm$ 0.16  & 6.37 $\pm$ 0.04           \\
NCANDA                & 0.41 $\pm$ 0.09 & 5.95 $\pm$ 0.06            \\
ADNI                        & \textbf{0.45 $\pm$ 0.03}  & \textbf{5.88 $\pm$ 0.04}            \\ \bottomrule
 \end{tabular}
 }
\end{table}

\subsection{Classification Results on ADNI} \label{classification_results}
After pretraining on the ADNI dataset, we evaluated the proposed SOE framework against the baseline models on ADNI NC vs. AD binary classification as a downstream task. As shown in Table \ref{combined_table}, we observe that our best SOE model with robustness regularizer (Section \ref{model_robustness}) outperforms all other compared models by $1.1$-$4.7\%$ in BACC, $1.9$-$3.7\%$ in AUC, and $1.3$-$6.2\%$ in F1 Score. Specifically, the best SOE model without the robustness regularizer performs on par with SimCLR model and marginally worse than the best MAE model. This aligns with our hypothesis that explicitly modeling the SO(3)-equivariance in the representation space is key to leveraging valuable geometric information of 3D MRIs compared to approaches that neglect geometry altogether. 

\begin{table}[t]
\centering
\caption{ADNI classification and age regression comparison with SOTA methods.}\vspace{-8pt}
\label{combined_table}
\resizebox{0.85\textwidth}{!}{
\begin{tabular}{p{3.6cm}w{c}{2.1cm}w{c}{2.1cm}w{c}{2.1cm}w{c}{2.1cm}w{c}{2.1cm}}
\toprule
 & \multicolumn{2}{c}{NC vs. AD Classification} & \multicolumn{2}{c}{NC Age Regression} \\ \cmidrule(r){2-3} \cmidrule(l){4-5}
& BACC\textsuperscript{$\uparrow$} & F1\textsuperscript{$\uparrow$} & R2\textsuperscript{$\uparrow$} & MAE\textsuperscript{$\downarrow$} \\ \midrule
No pretraining & 82.34 $\pm$ 4.24  & 78.15 $\pm$ 5.56 & 0.36 $\pm$ 0.16 & 6.37 $\pm$ 0.04 \\
AE & 81.63 $\pm$ 4.62  & 76.68 $\pm$ 6.37 & \textbf{0.46 $\pm$ 0.02} & 5.69 $\pm$ 0.00 \\
VAE & 83.14 $\pm$ 3.81  & 79.06 $\pm$ 5.03 & 0.45 $\pm$ 0.02 & 5.70 $\pm$ 0.00 \\
MAE & 84.69 $\pm$ 2.28 & 80.75 $\pm$ 3.69 & \textbf{0.46 $\pm$ 0.05} & \textbf{5.67 $\pm$ 0.02} \\
SimCLR & 83.94 $\pm$ 4.03 & 79.73 $\pm$ 4.47 & 0.42 $\pm$ 0.09 & 6.07 $\pm$ 0.11 \\
SOE w/o reg. & 83.84 $\pm$ 4.11 & 79.38 $\pm$ 5.58 & 0.45 $\pm$ 0.03 & 5.88 $\pm$ 0.04 \\
\midrule
Proposed Method (SOE) & \textbf{85.62 $\pm$ 2.80} & \textbf{81.78 $\pm$ 4.21} & 0.45 $\pm$ 0.03 & 5.88 $\pm$ 0.04 \\ \bottomrule
\end{tabular}
}
\end{table}

\subsection{Age Regression Results on ADNI} \label{age_regression}
Next, we evaluated the proposed representation learning framework with regards to the age prediction task on the ADNI NC cohort. Under the NC age regression column in Table \ref{combined_table}, we observe that this regression task is quite challenging overall. 
We observe marginal improvements in MAE and R2 coefficient of our best SOE model against the baseline pretrained models VAE and SimCLR. AE and MAE models marginally outperform our proposed model by less than $3\%$ in terms of MAE and R2. 

\subsection{Robustness Against Rotation}
\begin{figure}[t]
\centering
  \includegraphics[width=0.475\linewidth]{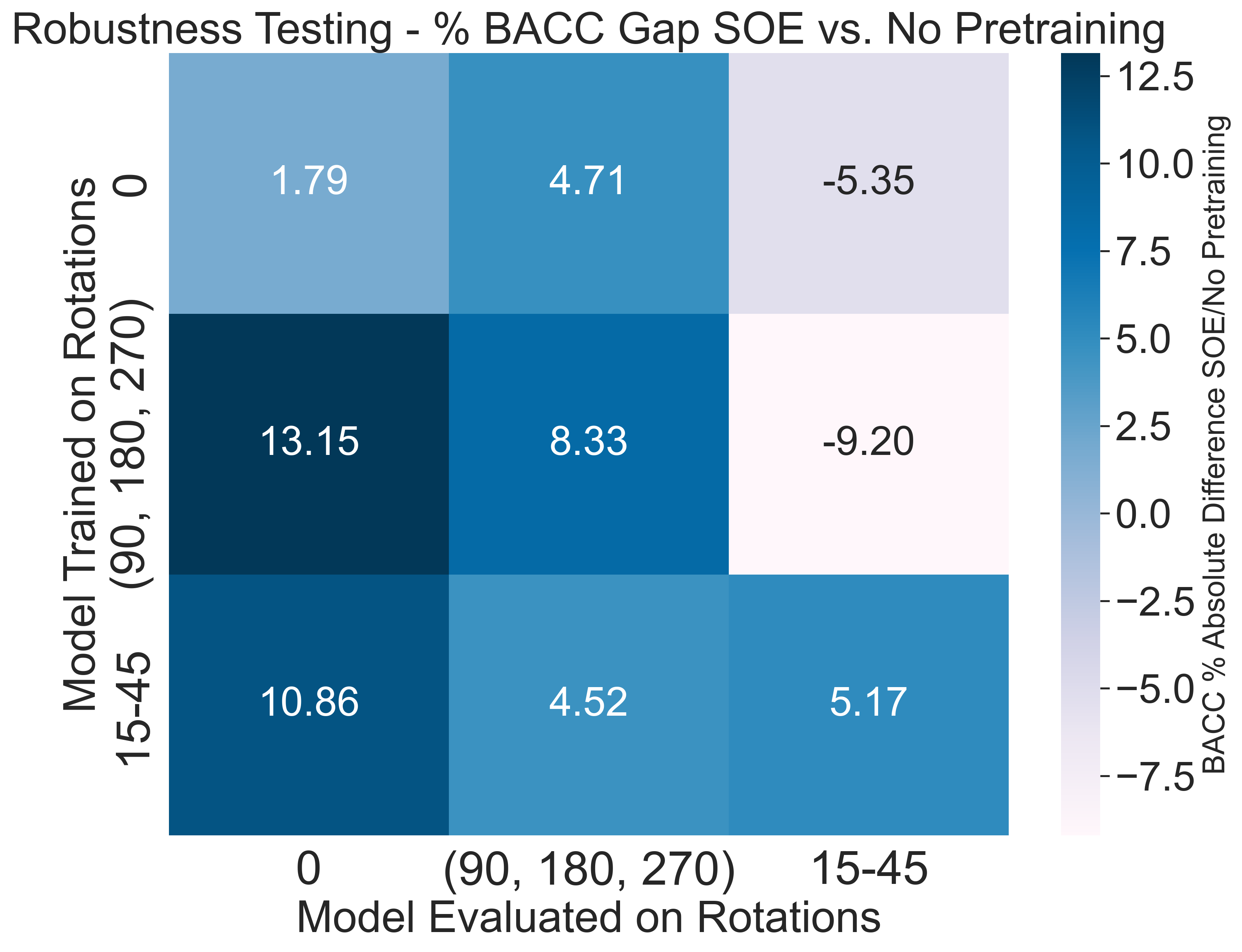}
    \hfill
  \includegraphics[width=0.45\linewidth]{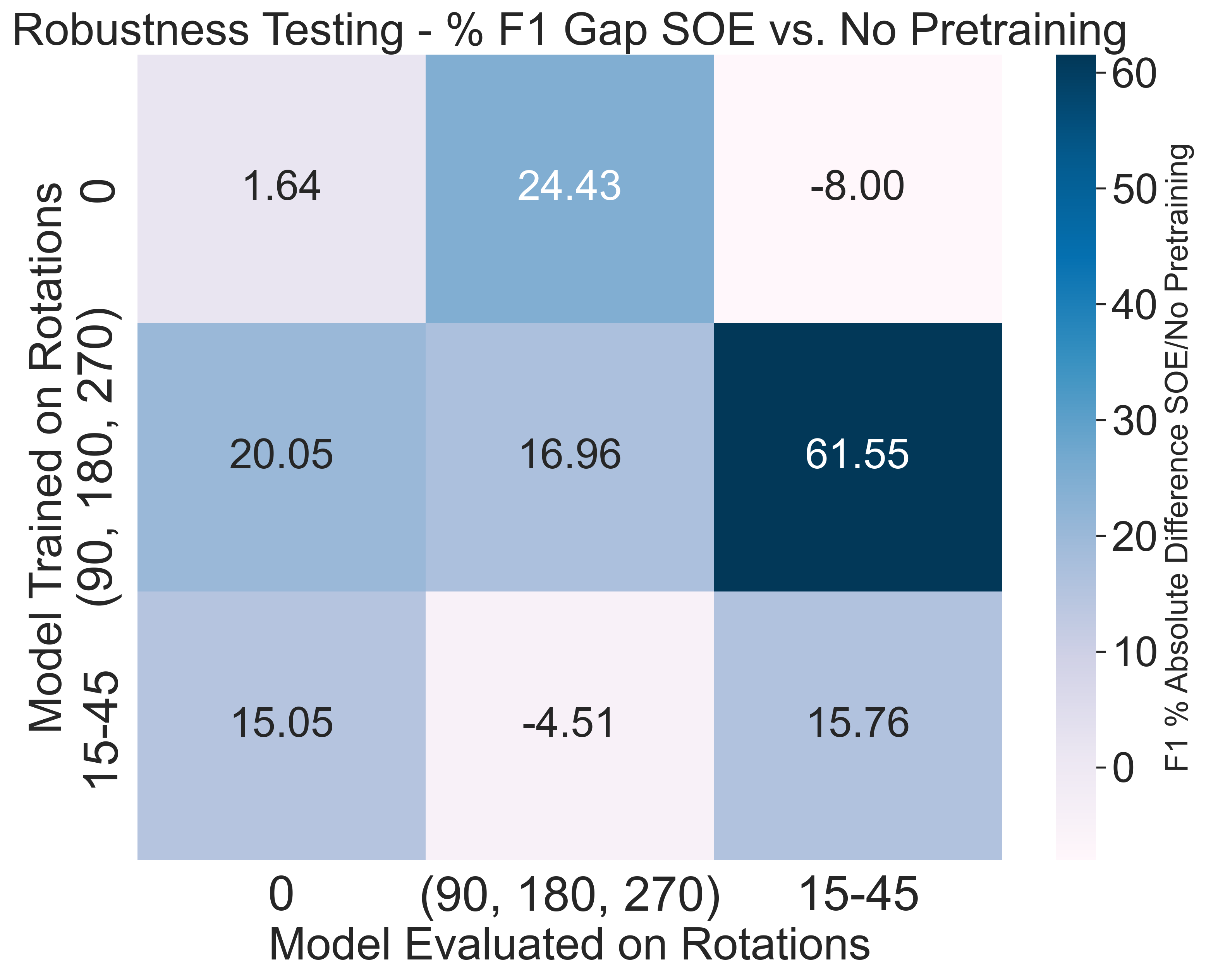}
\caption{ADNI Classification: Comparison of rotation augmented SOE vs. no pretraining model. Displayed are the \% increases in BACC (left) and F1 (right) from a model with no pretraining to a SOE pretrained model. The labels on each x axis shows the rotation degrees employed during training and the y axis rotation degrees employed during evaluation.}
\label{fig:robustness_grids}
\end{figure}

We further examined the robustness of the proposed SOE representation learning framework at different levels of rotations as data augmentation during downstream training. Specifically, we compare our best model against a model with no pretraining when trained on the ADNI classification task with different levels of rotation augmentation and evaluated on different levels of rotation. As seen in Fig.~\ref{fig:robustness_grids}, we observe increase in BACC and F1 for the SOE pretrained model when trained with rotation augmentations and evaluated on unrotated samples. Similarly, we see that the proposed SOE representation learning framework displays superior performance over a no pretraining approach in nearly all other scenarios. This confirms our expectation that when augmented with rotations during downstream training, a SO(3)-equivariant encoder can more easily adapt to rotations and implicitly learn rotational robustness than an untrained encoder. In theory, the classification head on a SO(3)-equivariant encoder would only have to learn SO(3)-invariance to ensure rotational robustness. 
\section{Conclusion}
We proposed a self-supervised representation learning framework, SOE, which models geometric SO(3)-equivariance in the feature representation space by leveraging the Vector Neuron Framework. By modeling SO(3)-equivariance, we move beyond conventional representation learning methods which neglect geometric information. Experiments show the generalizability of our representation learning approach across datasets and showed that SOE outperforms other SOTA representation learning methods on AD classification and is on par on age regression. Additionally, we illustrated the potential of SOE to facilitate more extreme rotations for data augmentation and its potential to increase model robustness against different levels of rotation. 
A direction for future work is to expand the notion of equivariance to more distortions that can be modeled as matrix transformations to enforce further geometric and mathematical coherence between the input and latent spaces.  

\subsubsection{Acknowledgements.} This work was supported by the HAI-Google Research Award, U.S. National Institute  (AG073104), the 2024 HAI Hoffman-Yee Grant, and the HAI-Google Cloud Credits Award.  

%
%

\bibliographystyle{splncs04}
\bibliography{paper}

\end{document}